\documentclass[a4paper,conference]{IEEEtran}
\ifCLASSINFOpdf
\else
\fi

\usepackage{graphicx}
\usepackage{comment}
\usepackage{amsmath,amssymb} % define this before the line numbering.
\usepackage{color}
\usepackage{cite}
\usepackage{subcaption}
\usepackage{authblk}
\usepackage{array}
\usepackage{booktabs}

\author[1, 3]{Kalun Ho}
\author[2]{Janis Keuper}
\author[1]{Franz-Josef Pfreundt}
\author[3]{Margret Keuper}
\affil[1]{Competence Center High Performance Computing, Fraunhofer ITWM, Kaiserslautern, Germany }
\affil[2]{Institute for Machine Learning and Analytics (IMLA), Offenburg University, Germany}
\affil[3]{Data and Web Science Group, University of Mannheim, Germany}
\date{}                     %% if you don't need date to appear
\begin{document}

\title{Learning Embeddings for Image Clustering: \\ An Empirical Study of Triplet Loss Approaches}


% --------------------------------------
% BEGIN: comment out this
% --------------------------------------
\iffalse

\author{\IEEEauthorblockN{Kalun Ho}
\IEEEauthorblockA{CC-HPC, Fraunhofer ITWM\\
Kaiserslautern, Germany\\
Email: kalun.ho@itwm.franhofer.de }
\and
\IEEEauthorblockN{Janis Keuper}
\IEEEauthorblockA{Institute for Machine Learning \\
and Analytics (IMLA)\\
Offenburg University, Germany\\
Email: keuper@imla.ai}
\and
\IEEEauthorblockN{Franz-Josef Pfreundt}
\IEEEauthorblockA{CC-HPC, Fraunhofer ITWM\\
Kaiserslautern, Germany\\
Email: kalun.ho@itwm.franhofer.de }
\and
\IEEEauthorblockN{Margret Keuper}
\IEEEauthorblockA{Data and Web Science Group\\
University of Mannheim, Germany\\
Email: keuper@uni-mannheim.de}
}
\fi
% --------------------------------------
% END: comment out this
% --------------------------------------

\maketitle


\begin{abstract}
%Learnt embedding features have been widely used in computer vision.
%Particularly for data clustering, learning discriminative features from a limited or noisy amount of %training data is a very challenging task.
%A popular approach is by comparing a query image with a positive and a negative sample.
%The positive pairs should be near each other while the negative pairs should lie far away from each other %in the embedding space.
%This can be achieved by optimizing the Triplet Loss.

In this work, we evaluate two different image clustering objectives, k-means clustering and correlation clustering, in the context of Triplet Loss induced feature space embeddings. % a graph-based clustering method based on embedding features learned from training data.
Specifically, we train a convolutional neural network to learn discriminative features by optimizing two popular versions of the Triplet Loss in order to study their clustering properties under the assumption of noisy labels.\\
Additionally, we propose a new, simple Triplet Loss formulation, which shows desirable properties with respect to formal clustering objectives and outperforms the existing methods. We evaluate all three Triplet loss formulations for K-means and correlation clustering on the CIFAR-10 image classification dataset.
%An extensive study uncovers some key findings when using graph-based clustering methods.

\end{abstract}


% \fi
\IEEEpeerreviewmaketitle

%%%%%%%%% BODY TEXT
\section{Introduction}
\label{sec:introduction}
When grouping data with missing or noisy labels, unsupervised approaches such as clustering are crucial.
%Clustering has been widely studied in unsupervised learning.
While fully supervised classification methods might result in a higher level of accuracy, they require a large amount of annotated data.  
%The advantage over supervised methods is obvious: while many focus on specific problems or tasks and thus their models are highly optimized to deliver best performance, they require a large amount of supervision.
In contrast, clustering approaches leverage the intrinsic data properties such as data density distributions or pairwise distances instead of annotations in order to group. % do not need any label data.
However, finding a suitable distance metric is essential.
One such a metric could be the Euclidean distance of two data points based on some features. For instance \textit{k-means} clustering is based on the Euclidean distance to some centroid. Thus, the properties of the features space play a crucial role. In computer vision, the simplest choice would be to use raw image pixel. 
However, this is often neither effective nor feasible for images with higher resolutions, and it shows poor generalization.
An effective way is to first reduce the dimension of the image (e.g. raw image pixels) into a lower dimensional space (embedding) with a non-linear mapping function.
This can be achieved using a convolutional neural network (CNN).
A popular loss function for learning such an embedding is the Triplet Loss~\cite{schroff2015facenet}, where three images are given such that the CNN learns to organize images of the same class closer to one another in the embedding space than images of different classes. 
Using the resulting embedding as features, one can run traditional clustering methods such as \textit{k-means} clustering as for example done in~\cite{xie2016unsupervised}. \textit{k-means} clustering assumes that the data points are evenly distributed around the cluster centroids. Furthermore, it is required that the number of clusters has to be specified beforehand. In many practical tasks, this specific scenario might be unrealistic because, for example, the number of objects to be grouped is simply unknown or because data points from the same class lie on a more complex manifold.
This motivates us to additionally consider a graph-based approach, where no data specific knowledge is required, i.e. the \emph{minimum cost multicut} problem, also known as \emph{correlation clustering}~\cite{chopra-1993,demaine-2006}.
%We we believe that this is more suitable for real-world problems, where the number of clusters are unknown.

\begin{figure}[t]
	\begin{center}
		\includegraphics[width=1.0\linewidth]{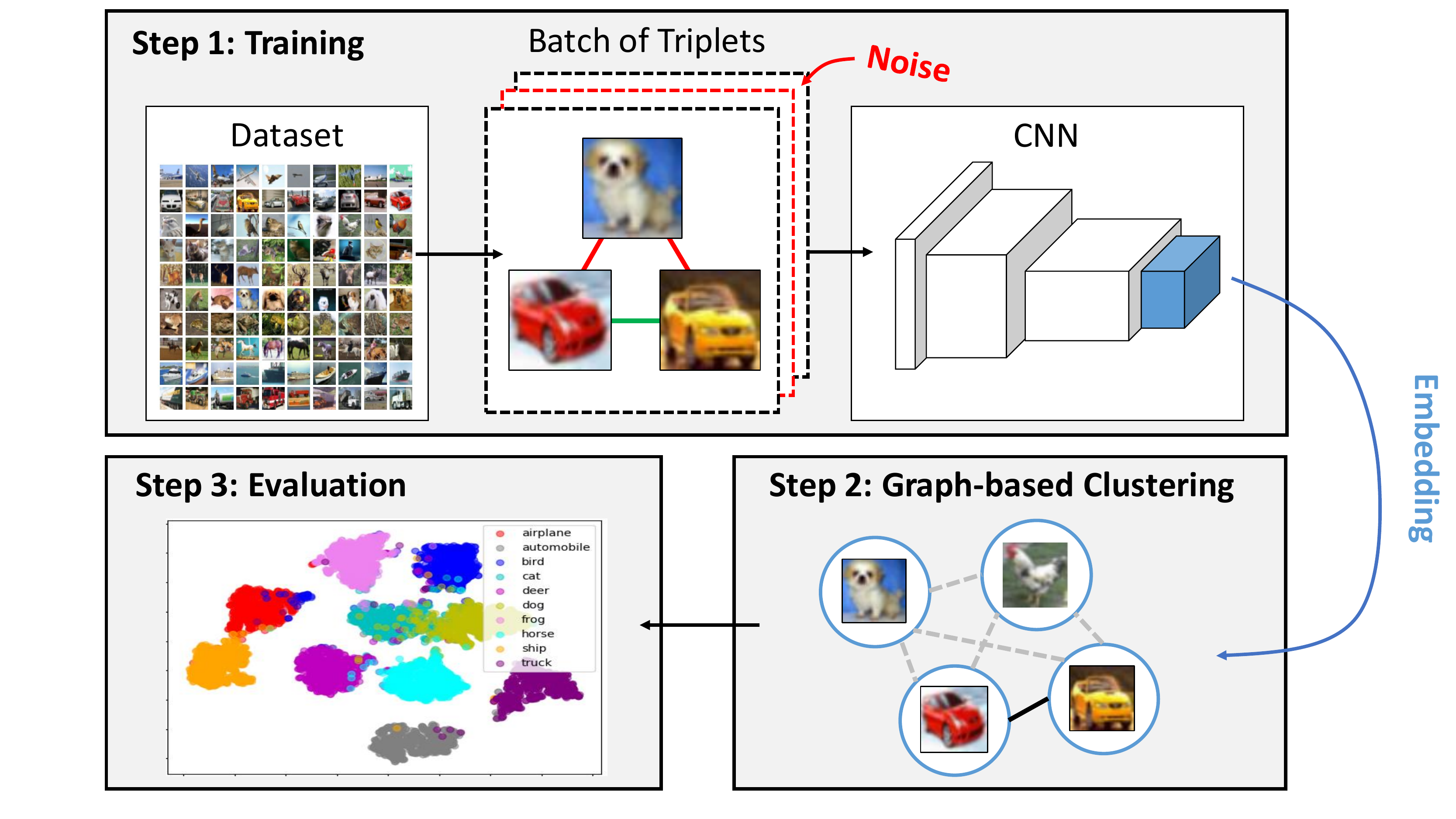}
	\end{center}
	\caption{Summary of the experiment setup in three steps: 1) dataset is trained using the Triplet Loss. Here, we add random noises (in red) by selecting wrong samples in the training data. 2) We cluster data using graph-based approach based on the learned embedding features (in blue). 3) Evaluation of clustering accuracy.}
	\label{fig:teaser}
    \vspace*{-3mm}
\end{figure}
In the context of these two common clustering techniques, we want to study the properties of embeddings resulting from two common variants of the Triplet Loss \cite{schroff2015facenet,zhang2016deep} and investigate their susceptibility to label noise in the training data. Additionally, we propose and study a third variant of the Triplet Loss, which shows promise in the context of both minimum cost multicuts as well as \textit{k-means} clustering and can be understood as a simplification of the loss proposed in \cite{zhang2016deep}. Both share the desired property to directly allow the extraction of pairwise cut probabilities between data points from the embedding space without an intermediate learning step. 
Specifically, we train a CNN on the CIFAR-10 image classification  dataset~\cite{krizhevsky2009learning} to learn discriminative features using the three variants of the Triplet Loss, where we apply noise to the training labels for positive and negative samples.
We evaluate the resulting embeddings by comparing the resulting clustering performance using minimum cost multicuts and \textit{k-means} clustering.
Figure~\ref{fig:teaser} illustrates our experimental setup.
\newpage

In summary, our contributions are:

\begin{itemize}
    \item We conduct a thorough study of the clustering behavior of two popular clustering approaches, \textit{k-means} and minimum cost multicuts, applied to learnt embedding spaces from three Triplet Loss formulations on the CIFAR-10~\cite{krizhevsky2009learning} dataset under a varying amount of label noise. %graph-based clustering approach based on learned features from a convolutional neural network (CNN).
    \item{Our study reveals that, while the traditional Triplet Loss~\cite{schroff2015facenet} is well suited for \textit{k-means} clustering, its performance drops under the looser assumptions made by minimum cost multicuts.}
    \item{We propose a simplification of the Triplet Loss from \cite{zhang2016deep}~\eqref{eq:triplet_improved}, which allows to directly compute the probability of two data points for belonging to disjoint components and is robust against noise in both clustering scenarios.}
    \item{Our proposed Triplet Loss variant outperforms both previous versions in terms of clustering performance and stability under label noise on the CIFAR-10 dataset.}
	%\item We propose two variations of the Triplet Loss and show, that the optimization parameter can be used directly to compute the decision boundary for the graph-based clustering
	%\item We further show, that the our proposed Triplet Loss~\eqref{eq:triplet_improved} is more stable against noise
	%\item We conduct an extensive study on graph-based clustering using embedding features trained on three variations of the Triplet Loss on CIFAR10~\cite{krizhevsky2009learning} dataset and compare the performance with \textit{k-means} clustering methods
\end{itemize}

%\newpage
\section{Related Work}
\label{sec:related}

\subsection{Clustering}
Many clustering approaches on computer vision problems are based on dimensionality reduction, where a non-linear mapping function is applied. 
One popular way is to use an autoencoder, where an input image is encoded into a embedding of lower dimension and then the decoder attempts to reconstruct its original.
The embedding is then used as feature space for the clustering methods: For instance, 
Xie et al.~\cite{xie2016unsupervised} first train an autoencoder and then use the same dataset and fine-tune it by training it again using a KL-divergence loss.
Another approach based on autoencoder is~\cite{ghasedi2017deep}, which uses the reconstruction loss along with relative entropy to jointly train the network.
Similar approaches can be found in~\cite{yang2017towards, tian2014learning, ji2017deep}.
Recently, generative models have been proposed for clustering tasks~\cite{mukherjee2019clustergan, ghasedi2019balanced}.
A large scale study on clustering is proposed by Caron et al.~\cite{caron2018deep}, which iteratively groups the features with a \textit{k-means} during the optimization.

\subsection{Correlation Clustering}
Correlation Clustering, also referred to as the \emph{minimum cost multicut problem}~\cite{chopra-1993,demaine-2006} is a popular choice when the number of clusters are unknown.
One such practical scenario is multiple object tracking, where pedestrians are tracked by just providing their detections~\cite{keuper2018motion, sheng2018heterogeneous, chu2019famnet, henschel2018fusion, henschel2019multiple}. Correlation clustering allows to group the data points based on pairwise cut probabilities without any cluster size bias and optimizes the number of clusters along with the data association. 
The crucial part there is to define or learn cut probabilities based on features.
For instance~\cite{tang2016multi} uses DeepMatching~\cite{weinzaepfel2013deepflow} as a similarity measure, followed by a logistic regression.
An alternative is to use features from embeddings learnt through a Siamese network~\cite{tang2017multiple}. 
The training is based on pairs of images where the network outputs a binary decision.%, e.g. same or different person.
This is closely related to our work, since we also want to obtain discriminative features by training a CNN with the Triplet Loss.
However, instead of a binary output as done in~\cite{tang2017multiple}, we want our network to learn an embedding of a fixed vector length, e.g. 32 dimensions.

\subsection{Deep Embedding Learning}
The main idea of learning embeddings is to attract similar data points to one another in a lower dimensional space while pushing dissimilar samples away from one another.
While the Contrastive Loss~\cite{hadsell2006dimensionality} fixes the positive and negative pairs by a fixed distance, it can be restrictive to variations in the embedding space~\cite{wu2017sampling}.
%To allow for more variations, the Triplet Loss~\cite{schroff2015facenet} forces all negative pairs to be further away than its positive pairs.
In contrast, the Triplet Loss~\cite{schroff2015facenet} captures the relative similarity of pairs of data points instead of absolute similarities.
It has been widely used for embedding learning~\cite{oh2016deep, zhuang2016fast, hermans2017defense, dong2018triplet}.
However, Zhang et al.~\cite{zhang2016deep} highlighted three major issues and thus proposed an \emph{Improved Triplet Loss} by enforcing intra- and intercluster constraints. 
We compare these two Triplet Loss formulations and propose a simplification of the formulation from Zhang et al., which outperforms both other formulations in practice.
\section{Study Setup}
\label{subsec:method}

In this section, we describe the setup of our study.
First, we introduce the architecture and the different variations of the Triplet Loss in Section~\ref{subsec:architecture} and \ref{subsec:loss}.
Then, in section~\ref{subsec:multicut}, we explain the correlation clustering method, also called the minimum cost multicut problem. Section~\ref{subsec:clustering_cifar10} describes the used dataset as well as the evaluation metric.

\subsection{CNN-Architecture}
\label{subsec:architecture}
We use AlexNet~\cite{krizhevsky2012imagenet} as a CNN-backbone as done in~\cite{caron2018deep}.
Furthermore, we also replaced the local response layers with batch normalization as done in~\cite{caron2018deep}. 
However, in order to reduce the feature dimensionality, we changed the size of the last two fully-connected layers from 4096 to 64 and 32, respectively.

%We include the details of our model architecture as well as the parameters in the supplementary materials.

% ----------------------------------------------------------

\subsection{Loss Function}
\label{subsec:loss}
Given any model architecture with trainable parameters $\theta$, we want to map an input image $x_i$ with a non-linear function $f_\theta: X \rightarrow Z$, with $f(x_i) \in \mathbb{R}^d$. 
In our case, $x_i$ is the image of the $i$-th sample from the dataset with a total number of $n$ samples, ${x_i \in X}_{i=1}^n$ and $d$ is the dimension of the embedding space.
$d$ is much smaller than the dimension of the input image. 
Given a set of three images, $x_i^a$, $x_j^n$ and $x_k^p$, the embedding features are learned by simply minimizing the Triplet Loss~\cite{schroff2015facenet} over the parameters $\theta$ of our deep neural network.
Here, the parameter $\alpha$ sets the margin of the similarity difference between the positive sample $x_k^p$ and negative sample $x_j^n$ and the anchor image:

\begin{equation}
\label{eq:triplet}
L_{triplet} = \sum_{i=1}^{n}[\|f(x_i^a) - f(x_i^p)\|^2 - \|f(x_i^a) - f(x_i^n\|^2) + \alpha]_{+}
\end{equation}
\\

\begin{figure}[t]
	\begin{center}
		\includegraphics[width=0.8\linewidth]{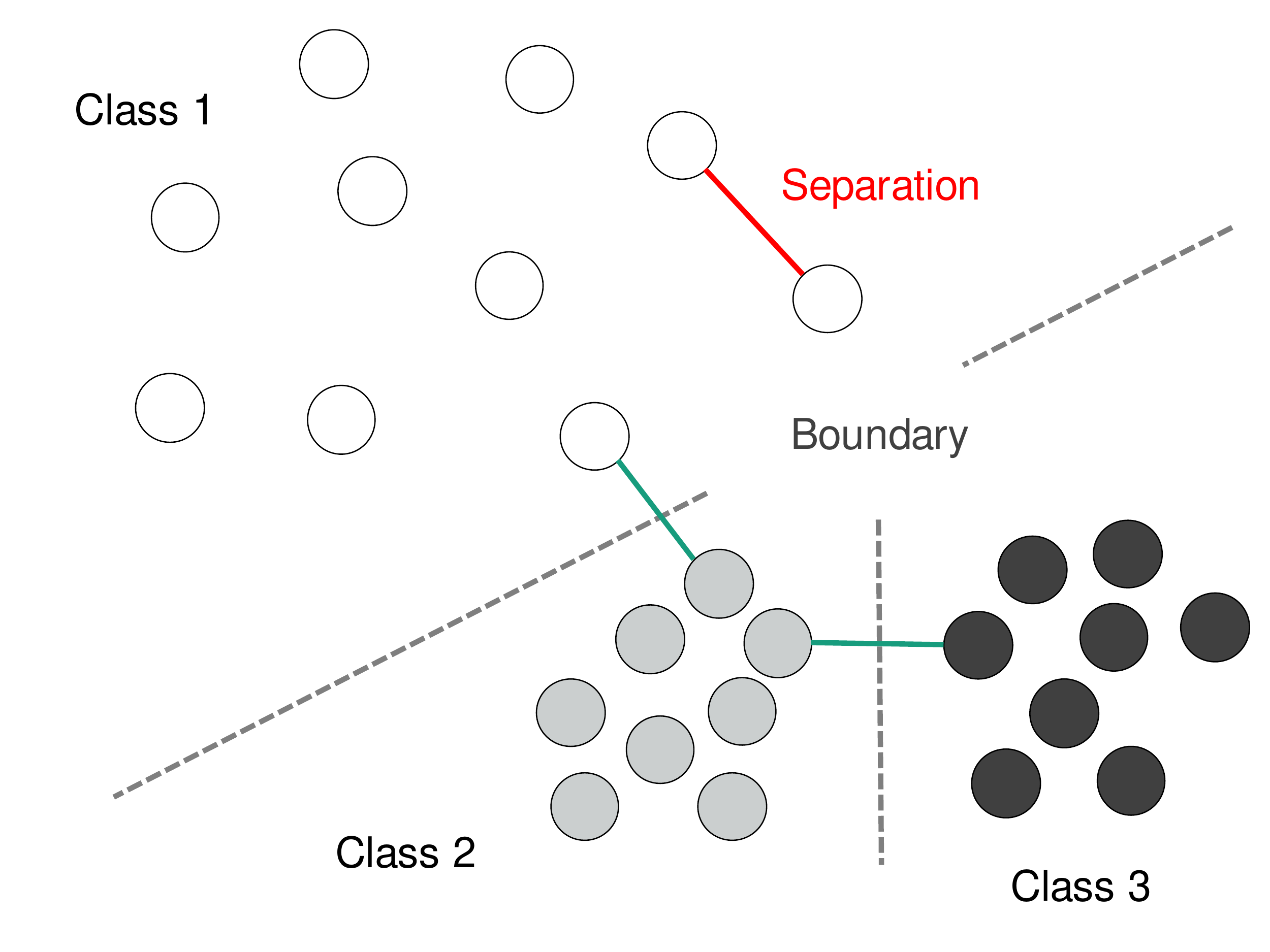}
	\end{center}
	\caption{Visualization of a possible data distribution trained with Triplet Loss~\cite{schroff2015facenet}: different intra-cluster distances over different classes make it impossible to learn one distance threshold at which data points should belong to different components. A logistic regression model could thus not learn cluster boundaries for graph partition. The correct decision  boundaries are marked by green lines while the red line shows a wrong separation of data. Our aim is to propose new Triplet Loss to stabilize such distances.}
	\label{fig:Fig1}
\end{figure}

Our approach is based on the assumption that embedding features, learned from the regular Triplet Loss~\eqref{eq:triplet} can produce high variances in inter- and intra-cluster distances, because it only considers relative differences between the distances of positive and negative pairs. This objective is suitable for k-means clustering. Yet, the attempt to learn whether two data points should belong to the same or to a different class from their pairwise distances (for example through a logistic regression model as in Section~\ref{subsec:clustering_cifar10}) might fail, when the intra- and inter cluster samples are equally far away.
This is shown in Figure~\ref{fig:Fig1} where the correct decision boundaries are marked by green lines.
In contrast, the red line, at the same Euclidean distance as the green lines, indicates a false separation of data.
This motivates us to consider losses that preserve the distance equally between the positive pairs during the optimization. Similar to \cite{zhang2016deep}, we therefore add an additional term to equation~\eqref{eq:triplet}, which we denote as \textit{Triplet Loss\_2}~\cite{zhang2016deep}:

\begin{equation}
\label{eq:triplet_add}
L_{triplet\_2} =  L_{triplet} + [\|f(x_i^a) - f(x_i^p)\|^2 - \beta]_{+}
\end{equation}
\\

The additive term in equation~\eqref{eq:triplet_add} introduces an additional parameter $\beta$, which sets the maximum distance between the positive pairs, e.g. the intra-cluster distance.
As the regular Triplet Loss only considers the distance \textit{difference} between the positive and negative pairs, set by the parameter $\alpha$, we propose a third loss function, which considers the absolute distance for the positive and negative pairs (instead of distance difference):

\begin{equation}
\label{eq:triplet_improved}
L_{triplet\_3} = [\alpha - \|f(x_i^a) - f(x_j^n)\|^2]_{+} + [\|f(x_i^a) - f(x_k^p)\|^2 - \beta]_{+}
\end{equation}
\\

We argue that this variant of the Triplet loss is more intuitive than Triplet Loss\_2 because it directly pushes positive pairs within a certain margin $\beta$ while driving negative pairs apart with a minimum distance $\alpha$. Within these margins, it still allows for varying distances, which is in contrast to for example Siamese approaches or the contrastive loss~\cite{hadsell2006dimensionality}.

\subsection{Minimum Cost Multicuts}
\label{subsec:multicut}
We assume, we are given an undirected graph {\it $G = (V, E)$}, where nodes $v\in V$ represent images and edges $e\in E$ encode their respective connectivity. 
Additionally, we are given real valued costs {\it $c: E \rightarrow \mathbb{R}$} defined on all edges which represent the node affinities. 
The goal is to determine \emph{edge} labels {\it $y: E \rightarrow \{0, 1\}$} defining a graph decomposition such that every partition of the graph corresponds to exactly one class. 
To infer such an edge labeling, we can solve instances of the minimum cost multicut problem with respect to the graph G and costs c, defined as follows \cite{chopra-1993,demaine-2006}:

\begin{align}
\min\limits_{y \in \{0, 1\}^E}
\sum\limits_{e \in E} c_e y_e
\label{eq:mc}
\end{align}

\begin{align}
s.t. \quad \forall C \in cycles(G) \quad \forall e \in C : y_e \leq \sum\limits_{e^\prime \in C\backslash\{e\}} y_{e^\prime}
\label{eq:cycle}
\end{align}
\\

The objective is to minimize \eqref{eq:mc} with respect to the assigned real valued costs of the edges and the corresponding cycle inequality constraint in Eq.~\eqref{eq:cycle}. 
The cycle inequality constraint ensures that the edge labeling $y$ induces a decomposition of $G$. In \cite{chopra-1993}, it was shown to be sufficient to enforce Eq.~\eqref{eq:cycle} on all \emph{chordless} cycles, i.e. all cycles.
Typically, if cut probabilities between pairs of nodes are available, the costs are computed using the \emph{logit} function $\text{logit}(p) = \log\frac{p}{1-p}$ to generate the real valued edge costs. 
With these costs set appropriately, the optimal solution of minimum cost multicut problems not only yields an optimal cluster assignment but also estimates the number of clusters automatically. 
Furthermore, this problem is able to generate small clusters and does not necessarily provide balanced sized clusters. 
\\

\subsubsection*{Optimization}
The minimum cost multicut problem~\eqref{eq:mc} is in the class of NP-hard~\cite{bansal-2004} and even APX-hard ~\cite{demaine-2006,hornakova-2017} problems. 
Nonetheless, instances have been solved within tight bounds, for example in \cite{andres-2012-globally} using a branch-and-cut approach. 
While this can be reasonably fast for some easier problem instances, it can take arbitrarily long for others. 
Thus, primal heuristics such as the one proposed in~\cite{keupericcv,node_agglom,Beier2016} or~\cite{CGC} are often employed in practice and show convincing results in various scenarios~\cite{keupericcv,tang2017multiple,insafutdinov2016deepercut}.

\newpage

\subsection{Pairwise Cut Probabilities for Multicuts}
\label{subsec:clustering_cifar10}

We formulate the following task as a minimum cost multicut problem:
each node in the graph $G$ represents an image $x_i$ that we want to assign to a class label.
The weight of the edges between two nodes represents the similarity between the two images.
The similarity is computed based on the Euclidean distance of in embedding feature using our CNN-model:

\begin{equation}
\label{eq:euclid}
d_{i, j} = \|f(x_i) - f(x_j)\|
\end{equation}

Once the similarity of two images is obtained, we seek to estimate their probability to belong to distinct classes. 
A decision function can be learned through a logistic regression using the label information, e.g. positive pairs are mapped to the value 0 while negative pairs are mapped to 1.

%We compute $T$ as shown in Figure~\ref{fig:Fig1} and then transform the value into a probability $p$. 
However, with the Triplet Loss\_2~\eqref{eq:triplet_add} and the proposed Triplet Loss\_3 in \eqref{eq:triplet_improved}, the distance threshold $\tau$ that decides whether two points should belong to different components (compare the distances in Figure~\ref{fig:Fig1}) can be automatically derived from the parameters $\alpha$ and $\beta$.
Since $\beta$ restricts the maximum distance of positive pairs to exactly $\beta=\alpha/2$, the distance threshold $\tau$ of the logistic function is computed as:

\begin{equation}
\label{eq:threshold}
\tau = \sqrt{(\alpha+\beta)/2}
\end{equation}

Note that this is not possible for the Triplet Loss from Eq.~\eqref{eq:triplet} because it only considers relative distances.

A complete graph is built and the clusters are obtained by optimizing equation~\eqref{eq:mc} using \cite{keupericcv}.
We use clustering accuracy (ACC) as evaluation metric, where the best map between predicted clusters and true label is found.

%%%%%%%%% BODY TEXT
\section{Experiments and Results on CIFAR-10}
\label{sec:results}

In this Section, we present the experiments and the results of our study.
CIFAR-10~\cite{krizhevsky2009learning} is a popular dataset for image classification tasks. 
It contains 50.000 train and 10.000 test data samples of tiny images (32px x 32px). 
Each sample is assigned a label that belongs to one of the ten classes: \textit{airplane}, \textit{automobile}, \textit{bird}, \textit{cat}, \textit{deer}, \textit{dog}, \textit{frog}, \textit{horse}, \textit{ship}, and \textit{truck}.
All models are trained for 100 epochs with a batch size of 100 and a learning rate of 0.001 using AdamOptimizer~\cite{kingma2014adam}.
First, we compare the performance of the minimum cost multicuts and k-means clustering using different Triplet Losses in Section~\ref{subsec:result_acc}.
Then, we present some insights related to inter- and intra-cluster distances in Section~\ref{subsec:result_dist}.
In Section~\ref{subsec:noise}, we present our study on the feature learning under label noise. The results are shown in Table~\ref{tab:table1}.
In Section~\ref{subsec:qualitative}, we present some qualitative results.

\begin{figure}[t]
	\begin{center}
		\includegraphics[width=1.0\linewidth]{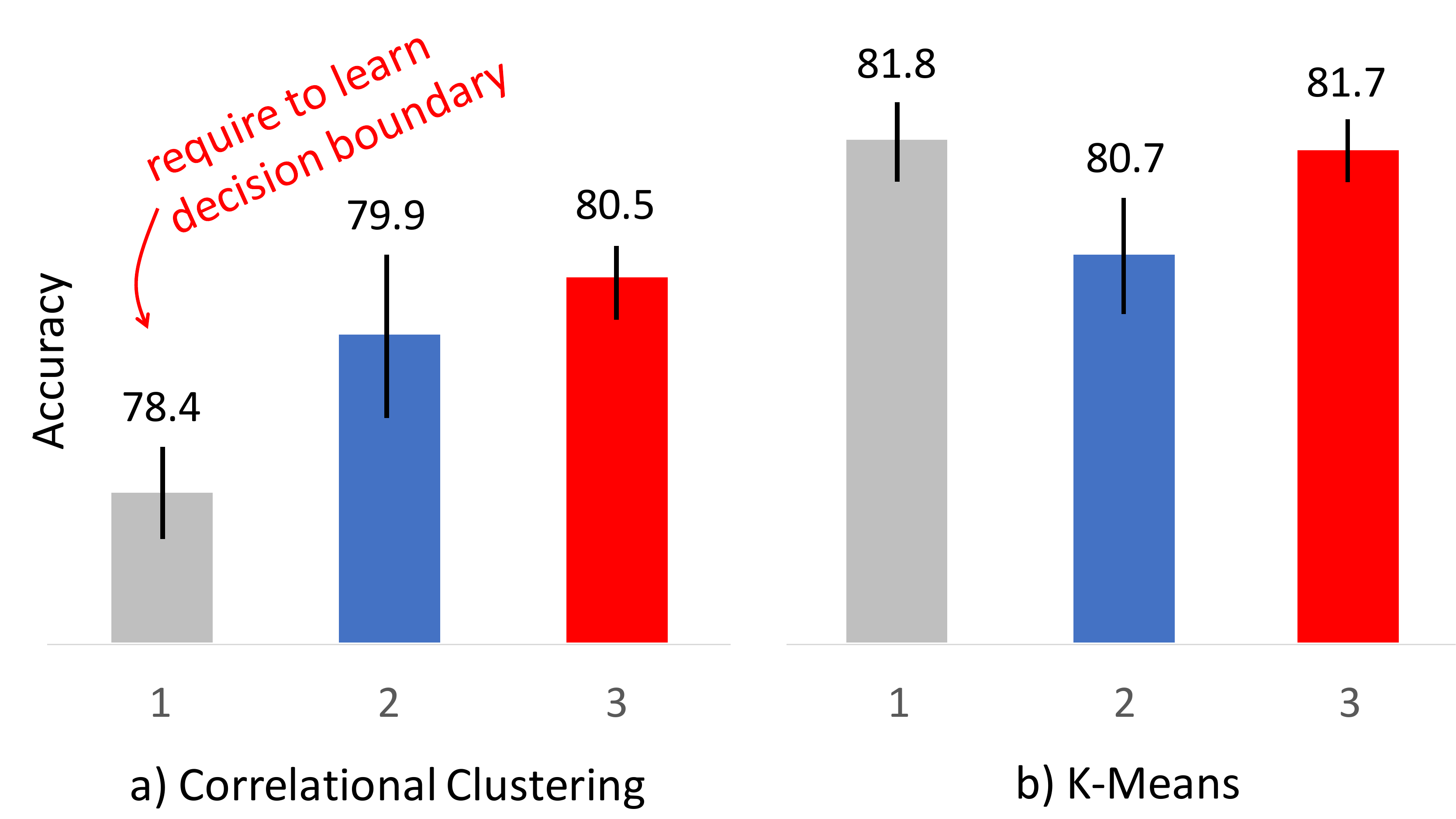}
	\end{center}
	\caption{Comparison of three different losses on clustering performance. The numbers on the x-axis represent the triplet loss type while the y-axis shows the average cluster accuracy on five runs. The black line indicates the standard deviation. Triplet Loss\_2~\eqref{eq:triplet_add} and our Triplet Loss\_3~\eqref{eq:triplet_improved} perform better than the regular Triplet Loss for minimum cost multicuts. However, k-means consistently achieves better results given the fact that the parameter $k$ is set correctly (shown in right). For k-means, the Triplet Loss\_2 shows worse performance than the regular one~\cite{schroff2015facenet}, while the proposed, simpler version, Triplet Loss\_3, performs best in both scenarios. }
	\label{fig:cifar10}
    \vspace*{-3mm}
\end{figure}
\iffalse
\begin{figure}[!htb]
\small
\begin{tabular}{c}
    \includegraphics[width=0.8\linewidth]{Fig_2_dist_1.pdf}\\
	a) Triplet Loss\_1 \eqref{eq:triplet}\\
	\includegraphics[width=0.8\linewidth]{Fig_2_dist_2.pdf}\\
	b) Triplet Loss\_2 \eqref{eq:triplet_add}\\
		\includegraphics[width=0.8\linewidth]{Fig_2_dist_3.pdf}\\
	c) Triplet Loss\_3 \eqref{eq:triplet_improved}
	\end{tabular}
    \caption{Euclidean inter and intra cluster distance trained with all three Triplet Loss variants (equation~\eqref{eq:triplet}, \eqref{eq:triplet_add} and \eqref{eq:triplet_improved}): the average distance shows significant variance when trained with a) Triplet Loss\_1~\eqref{eq:triplet}, especially for the class \emph{cat} and \emph{dog}. The overlap of the distances prevents the logistic regression model to learn the cluster boundaries. Both other variants, b) and c), produce more consistent and stable results, thus higher clustering accuracy.}
	\label{fig:dist_compare}
    \vspace*{5mm}
\end{figure}
\fi
\begin{figure*}[!htb]
	\begin{center}
		\includegraphics[width=0.99\linewidth]{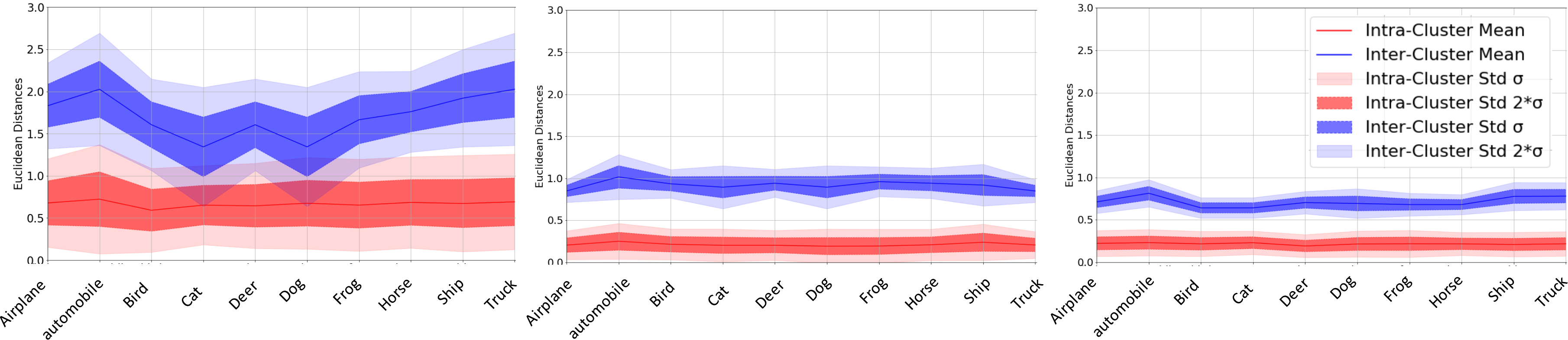}
	\end{center}	\vspace{-0.3cm}
	\small
	\hspace{1.8cm}a) Triplet Loss\_1 \eqref{eq:triplet}\hspace{3.5cm}b) Triplet Loss\_2 \eqref{eq:triplet_add}\hspace{3.5cm}c) Triplet Loss\_3 \eqref{eq:triplet_improved}
    \caption{Euclidean inter and intra cluster distance trained with all three Triplet Loss variants (equation~\eqref{eq:triplet}, \eqref{eq:triplet_add} and \eqref{eq:triplet_improved}): the average distance shows significant variance when trained with a) Triplet Loss\_1~\eqref{eq:triplet}, especially for the class \emph{cat} and \emph{dog}. The overlap of the distances prevents the logistic regression model to learn the cluster boundaries. Both other variants, b) and c), produce more consistent and stable results, thus higher clustering accuracy.}
	\label{fig:dist_compare}
    \vspace*{7mm}
\end{figure*}
\subsection{Evaluation of Cluster Accuracy}
\label{subsec:result_acc}

We compare the CNN models that are optimized with the three different losses. 
Furthermore, we also run k-means clustering on the same embedding features with $k$ correctly set to $k=10$ (thus introducing external knowledge to the clustering process).
For Tripet Loss~\eqref{eq:triplet}, an additional regression model is trained using the label information in order to estimate the threshold, while for~\eqref{eq:triplet_add} and \eqref{eq:triplet_improved}, the threshold is computed directly from the optimization parameters using equation~\eqref{eq:threshold}.
All experiments are executed five times with different random seeds and we report the average number over the clustering accuracy.
\\

\subsubsection*{Results} 
Figure~\ref{fig:cifar10} shows the clustering accuracy for different variants of the Triplet Loss.
First, we observe that Triplet Loss\_2~\eqref{eq:triplet_add} and Triplet Loss\_3 ~\eqref{eq:triplet_improved} outperform~\eqref{eq:triplet} on multicut clustering (left).
However, K-Means (right) performs better on all our experiments given the fact that $k=10$ is known.
The highest performance is achieved when we train the CNN-model with the Triplet Loss\_3~\eqref{eq:triplet_improved}, where the average accuracy is 80.5\% (red). For k-means, the Triplet Loss\_2 shows worse performance than the regular one~\cite{schroff2015facenet}, while the proposed, simpler version, Triplet Loss\_3, performs best in both scenarios. % cluster acc vs. margin

% -------------------------------------------------------------------

\subsection{Inter- and Intra-cluster distances}
\label{subsec:result_dist}

Figure~\ref{fig:dist_compare} compares the cluster distances of the samples of all three Triplet Losses.
The red curve represents the average pairwise distances of the samples within a cluster while the blue curve shows the average distances of one cluster to its nearest cluster.
Furthermore, the color range represents the standard deviation $\sigma$ and $2\sigma$ of the cluster distances.
As illustrated in Figure~\ref{fig:dist_compare} a), both distances show significant variances when trained with the regular Triplet Loss~\eqref{eq:triplet}. 
Furthermore, class \textit{cat} and \textit{dog} show a significant overlap in their inter- and intra cluster distances, which prevents the logistic regression from setting the right decision boundary as explained in Figure~\ref{fig:Fig1}.
In contrast, Triplet Loss\_2~
\eqref{eq:triplet_add} and Triplet Loss\_3~\eqref{eq:triplet_improved} produce more consistent and stable results, and produce higher clustering accuracy (see Figure~\ref{fig:cifar10}). % Intra cluster variance

\begin{table*}[!htbp]
\centering
\begin{tabular}{c | c c c c c | c c c c c | c c c c c}
\toprule
{} &  \multicolumn{15}{c}{\textbf{Minimum Cost Multicuts}}\\
{} &  \multicolumn{5}{c}{Triplet Loss~\eqref{eq:triplet}} & \multicolumn{5}{c}{Triplet Loss~\eqref{eq:triplet_add}} & \multicolumn{5}{c}{Triplet Loss~\eqref{eq:triplet_improved}} \\
\midrule
Noise \% & random & 7\% & 5\% & 2\% & 0\% & random & 7\% & 5\% & 2\% & 0\% & random & 7\% & 5\% & 2\% & 0\% \\
\midrule
{20\%} & 73.85 & 73.64 & 74.43 & 76.28 & 75.18 & 60.13 & 65.61 & 68.05 & 67.26 & 64.39 & 75.23 & 74.39 & 75.00 & 72.80 & 75.27\\ 
{10\%} & 77.90 & 77.02 & 76.93 & 78.04 & 77.87 & 74.23 & 75.64 & 75.23 & 74.75 & 75.48 & 77.10 & 76.53 & 76.47 & 76.06 & 76.29\\ 
{5\%} & 78.65 & 76.73 & 78.30 & 77.73 & 78.87 & 75.62 & 76.27 & 76.06 & 75.80 & 76.63 & 77.07 & 78.94 & 80.16 & 77.44 & 77.75\\ 
{0\%} & 78.14 & 77.76 & 78.31 & 78.56 & \textbf{78.44} & 77.07 & 77.46 & 77.47 & 80.41 & \textbf{79.96} & 80.25 & 80.38 & 80.65 & 80.42 & \textbf{80.51}\\ 
\bottomrule
{} &  \multicolumn{15}{c}{\textbf{k-means Clustering}}\\
{} &  \multicolumn{5}{c}{Triplet Loss~\eqref{eq:triplet}} & \multicolumn{5}{c}{Triplet Loss~\eqref{eq:triplet_add}} & \multicolumn{5}{c}{Triplet Loss~\eqref{eq:triplet_improved}} \\
\midrule
Noise \% & random & 7\% & 5\% & 2\% & 0\% & random & 7\% & 5\% & 2\% & 0\% & random & 7\% & 5\% & 2\% & 0\% \\
\midrule
{20\%} & 79.83 & 80.26 & 80.29 & 80.60 & 80.55 & 71.71 & 74.70 & 74.86 & 74.62 & 74.61 & 79.66 & 80.54 & 77.83 & 78.20 & 80.53 \\
{10\%} & 81.55 & 80.95 & 81.10 & 81.53 & 81.21 & 75.33 & 75.80 & 75.89 & 75.54 & 76.48 & 81.93 & 81.78 & 80.59 & 80.45 & 82.06 \\
{5\%} & 81.62 & 80.82 & 81.27 & 81.19 & 81.68 & 76.79 & 77.01 & 76.14 & 76.97 & 76.93 & 82.07 & 82.11 & 81.98 & 81.90 & 82.08 \\
{0\%} & 81.59 & 81.33 & 81.53 & 81.51 & \textbf{81.82} & 80.25 & 80.80 & 81.11 & 81.39 & \textbf{80.73} & 82.15 & 82.14 & 81.93 & 81.64 & \textbf{81.72} \\
\bottomrule

\end{tabular}
\caption{Evaluation of clustering accuracy using \textbf{minimum cost multicuts} (top) and \textbf{k-means} (bottom) based on the embedding features of the CNN-model trained on three different Triplet Loss variants. The average accuracies of five runs are reported in \%. Furthermore, the numbers in bold are the reported numbers from Figure~\ref{fig:cifar10}. The x-axis shows the amount of label noise on the negative samples up to 7\%, while y-axis shows the amount of label noise on the positive samples up to 20\%, respectively. The column \textit{random} selects the negative sample without employing the labels. On the CIFAR-10 dataset, the chances are 90.0\% to retrieve correct negative samples. }
\label{tab:table1}
\vspace*{5mm}

\end{table*}
\begin{figure*}[!h]
\centering
\begin{tabular}{lclc}
%\subfloat[]{
&Minimum Cost Multicuts && K-means\\
 \rotatebox{90}{$\quad$noisy positive labels}&\includegraphics[width=0.45\linewidth]{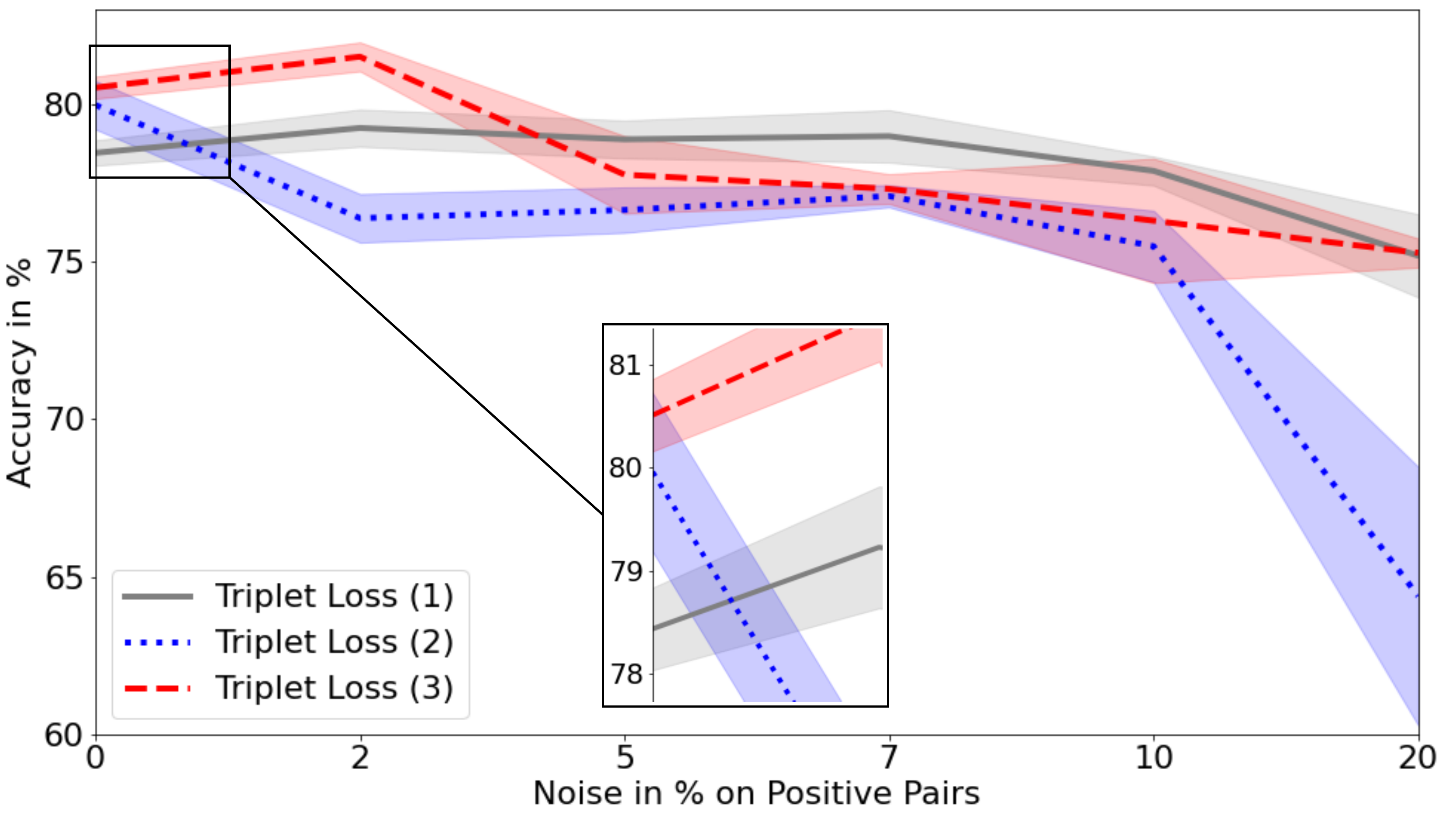}&
%}
%\subfloat[]{
  &\includegraphics[width=0.45\linewidth]{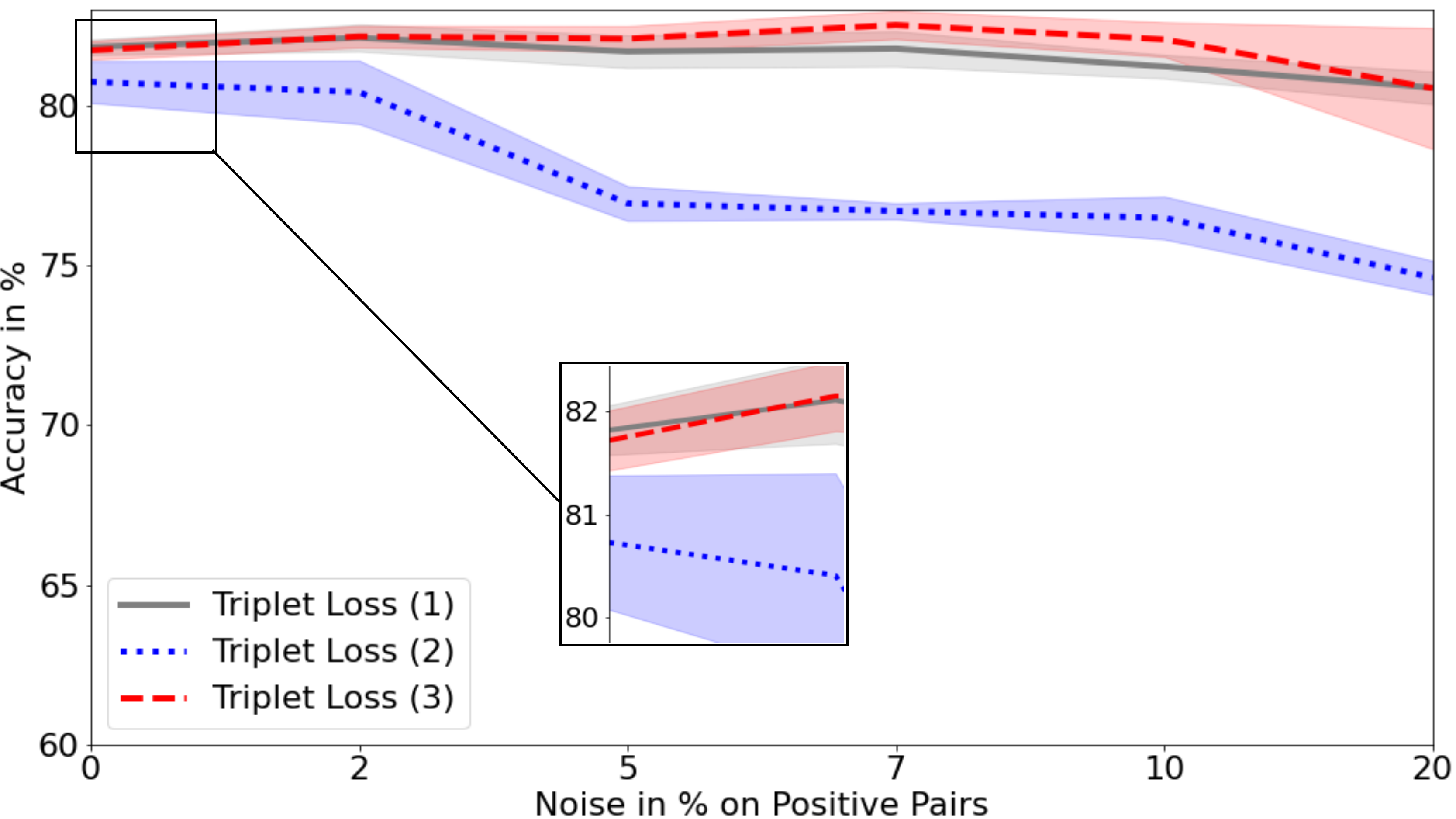}\\
 
%}
%\hspace{0mm}
%\subfloat[]{
  \rotatebox{90}{$\quad$ noisy negative labels}&\includegraphics[width=0.45\linewidth]{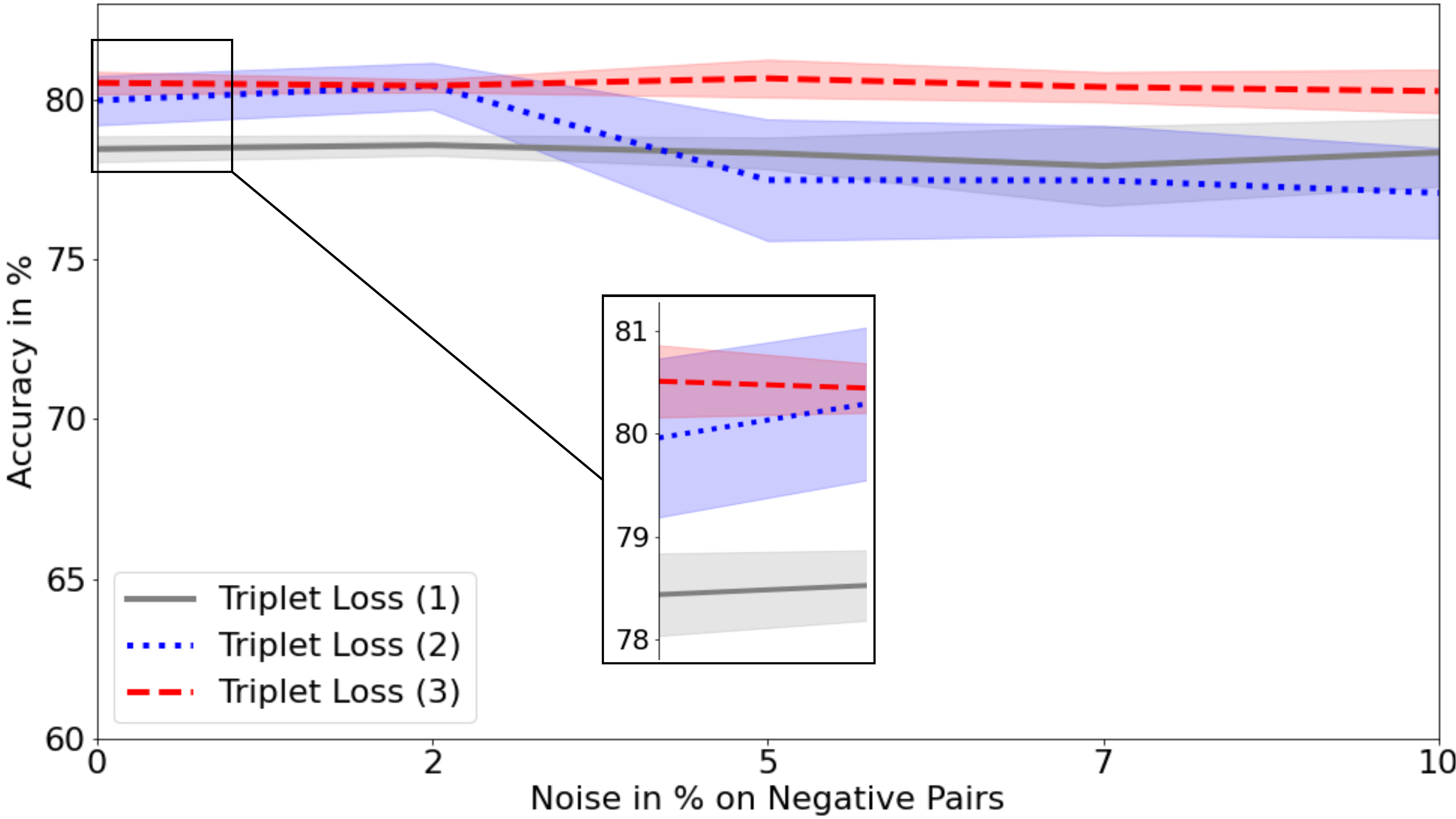}&
%}
%\subfloat[]{
  &\includegraphics[width=0.45\linewidth]{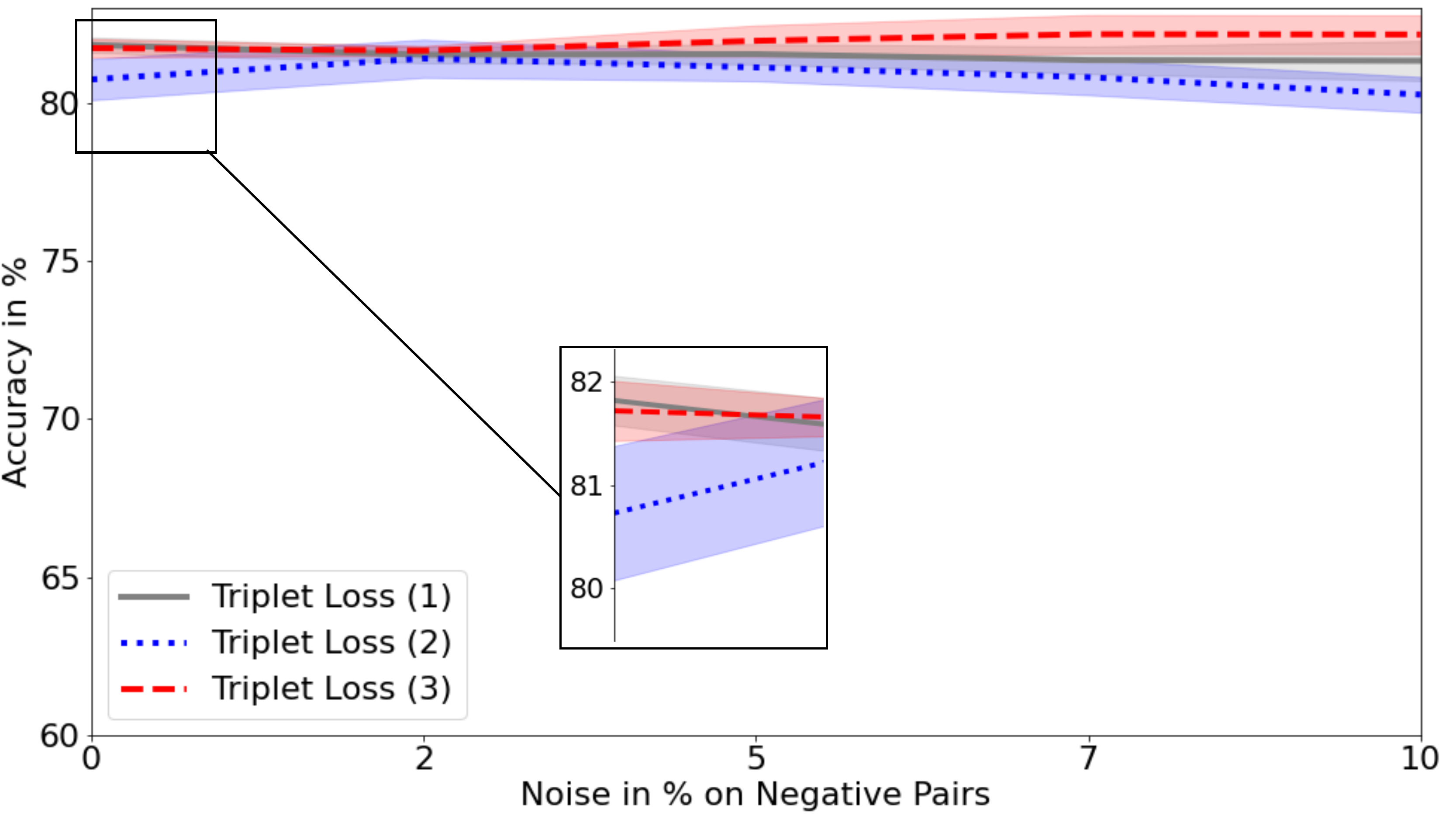}\\
%}
%\hspace{0mm}
%\subfloat[]{
  \rotatebox{90}{$\quad$ noise of pos. and neg.  labels}&\includegraphics[width=0.45\linewidth]{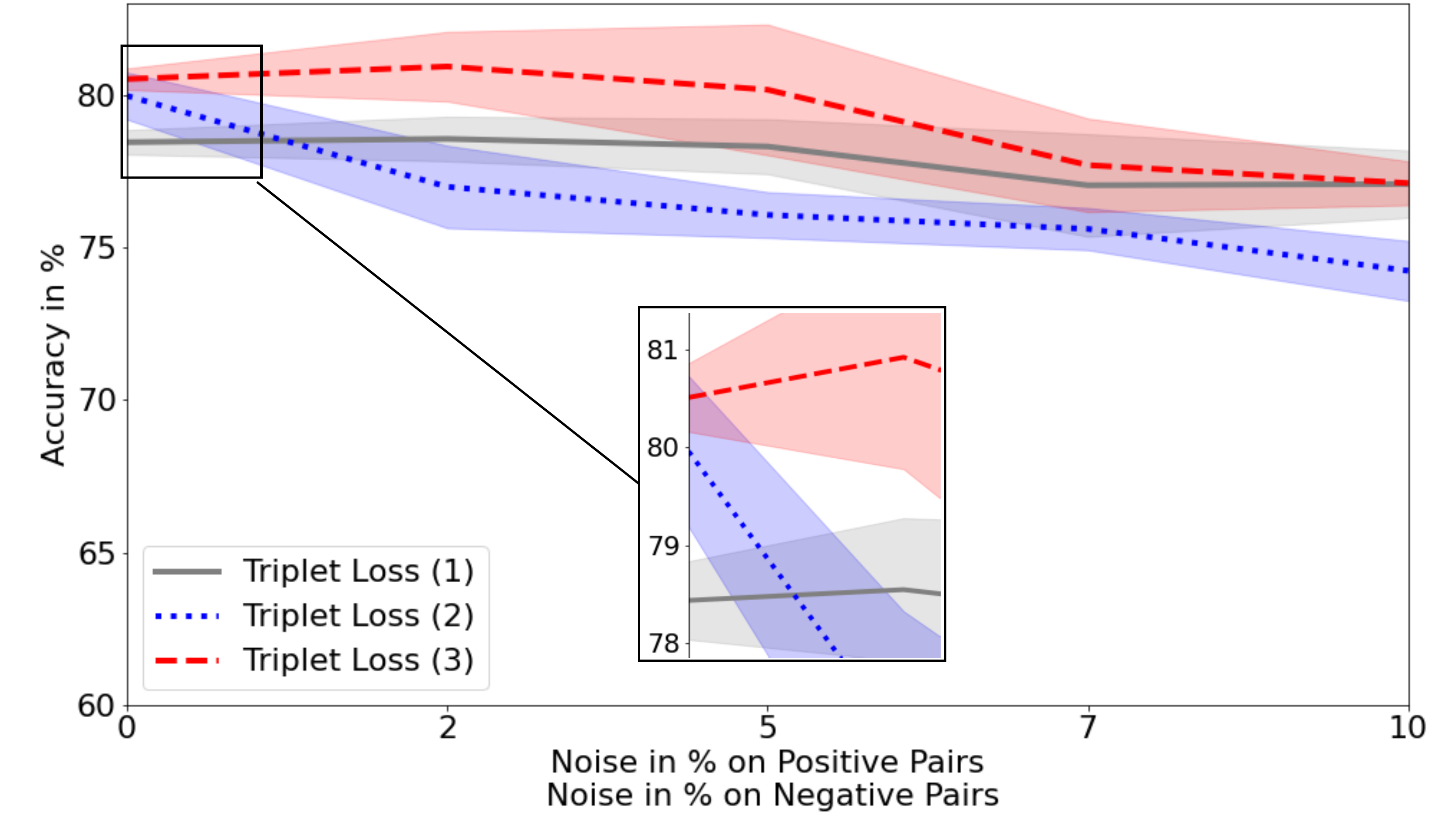}&
%}
%\subfloat[]{
  &\includegraphics[width=0.46\linewidth]{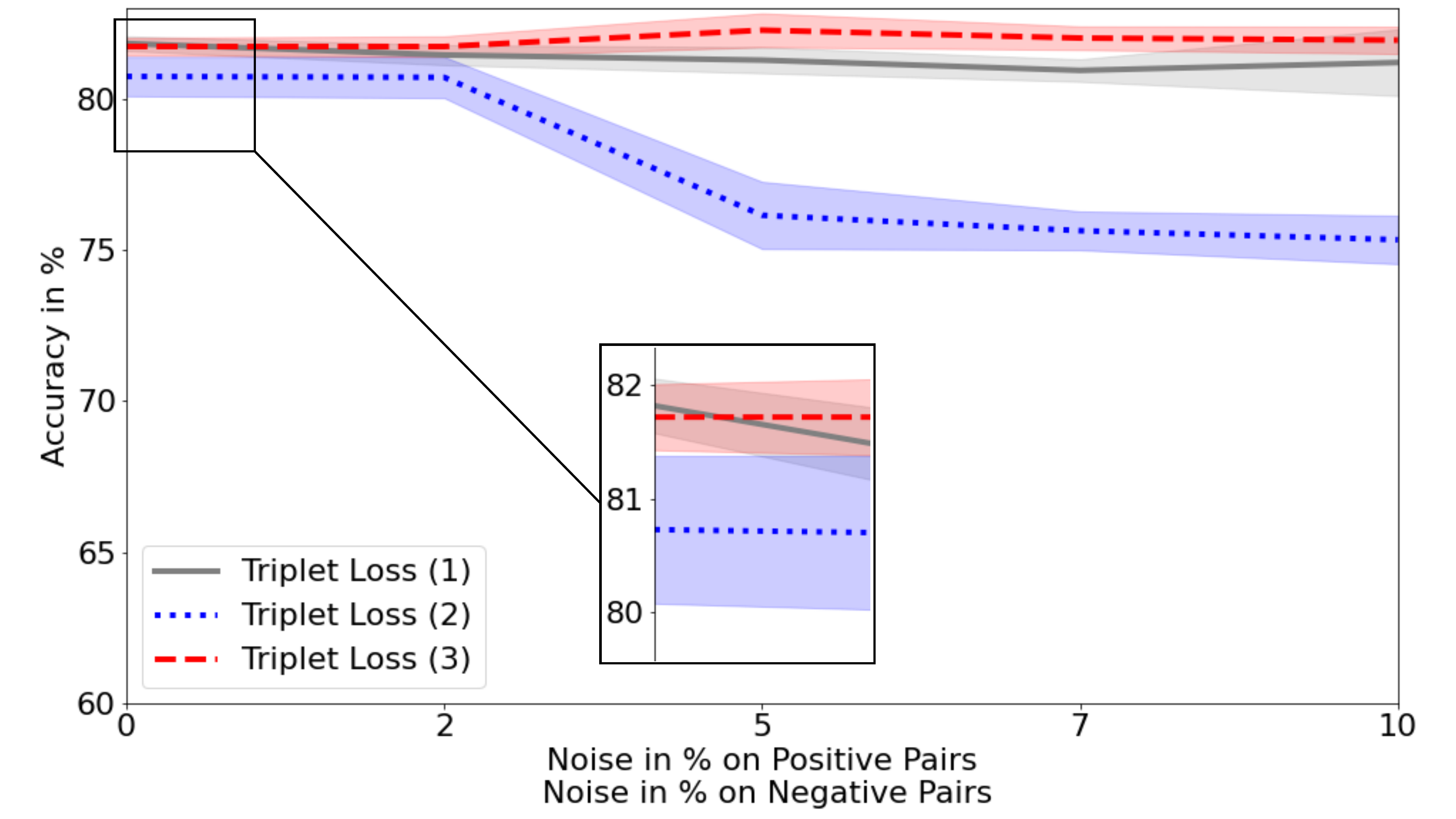}
%}
\end{tabular}
 \caption{Average cluster accuracy against the  percentage  of  noise,  that  is  applied  to  the  sampling. Noises on positive and negative pairs are applied on first and second row, respectively while the last row is evaluated on an equal amount of label noise of both, positive and negative pairs. The first column shows the results of the minimum cost multicuts while the second column employs k-means. The amount of noise (i.e. wrong pairs) is indicated in x-axis.}
\label{fig:acc_noise}
\end{figure*}
\subsection{Triplet Loss with Label Noise}

%\begin{figure*}[!h]
%	\begin{center}
%		\includegraphics[width=1.0\linewidth]{Fig_3_all.pdf}
%	\end{center}
%    \caption{Average cluster accuracy against the  percentage  of  noise,  that  is  applied  to  the  sampling. Noises on positive and negative pairs are applied on a) and b), respectively while c) is evaluated on both, positive and negative pairs. The amount of noise (e.g. wrong pairs) is indicated in x-axis. The experiments are all conducted on correlational clustering, except for d), which is marked in red (bottom right): here, we evaluated the k-means performance against noise.}
%	\label{fig:acc_noise}
%    \vspace*{5mm}
%\end{figure*}

\label{subsec:noise}

\begin{figure*}[!h]
	\begin{center}
		\includegraphics[width=1.0\linewidth]{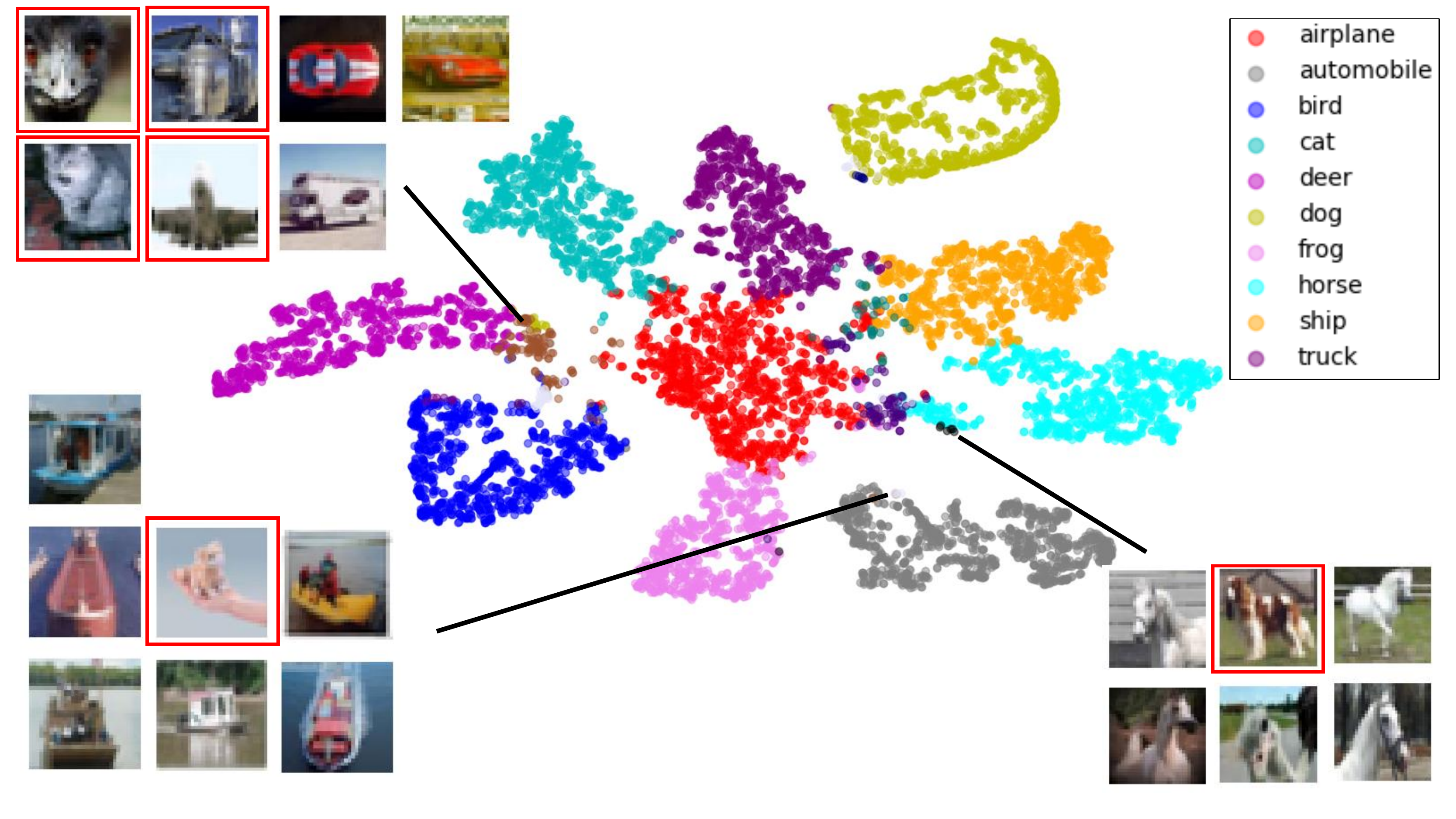}
	\end{center}
    \caption{TSNE-Visualization of the clusters on CIFAR-10 using a multicut approach. In this example, the cluster accuracy is 80.27\% trained on a CNN-model with the Triplet Loss~\eqref{eq:triplet_improved} and the total number of clusters is 44. 
    The color represents the 10 largest clusters found. There are 33 small clusters ($<10$ items), for instance bottom left and right are two clusters shown containing 7 images each. False positives within the clusters are marked as red. The cluster on the upper left corner contains images from bird, cat, automobile (x3), planes and trucks.} 
    %\vspace*{2mm}
	\label{fig:tsne}
\end{figure*}
In this experiment, we investigate the sensitivity of the different Triplet Losses towards label noise. 
Specifically, we randomly select \textit{wrong} triplets during the training process of our CNN-model and evaluate the clustering performance based on the embedding features, trained on all three loss variants from Subsection~\ref{subsec:loss}.
The experiments were conducted repeatedly five times with different seeds and we report the mean cluster accuracies in \%.
The parameters $\alpha$ and $\beta$ are fixed to $0.8$ and $0.4$, respectively.

\subsubsection*{Results}
Table~\ref{tab:table1} shows our complete evaluation with various setups:
we applied different amounts of label noise on triplets for positive and negative pairs.
The x-axis represents the noise for the negatives while the y-axis indicates noise on positive pairs within the triplet.
These \textit{wrong pairs} are retrieved randomly.
All results are reported as average clustering accuracy over five runs of training.
In Table~\ref{tab:table1} top, we present the performance using minimum cost multicuts while the bottom rows show the results of \textit{k-means} clustering
Note that \textit{k-means} requires to specify the number of clusters $k$, beforehand (on CIFAR-10, we know $k=10$), while minimum cost multicuts do not require this dataset specific knowledge.
Without noise added, considering the CNN-model trained with the same loss function, \textit{k-means} seems more stable against noise and outperforms the correlational clustering on average by 1-2\% and the highest clustering accuracy when no noises are added. In Figure~\ref{fig:acc_noise}, we give a more detailed analysis of these results.
\\

Remark: \textit{When sampling the triplets randomly on balanced dataset with $k$-clusters, the chance to get a true positive and true negative pair is $\frac{1}{k}$ and $\frac{k-1}{k}$ respectively}.
\newpage

Figure~\ref{fig:acc_noise} shows the clustering accuracy against the percentage of noise that is applied to the sampling. In Figure~\ref{fig:acc_noise}, top left, we only add noise to the positive pairs while selecting correct negative samples and evaluate using minimum cost multicuts, i.e. without introducing knowledge on the number of classes.
Our first observation is that Triplet Loss\_2~\eqref{eq:triplet_add} is the most sensitive to noise among all three loss variants.
This is shown in Figure~\ref{fig:acc_noise} in blue. 
The regular Triplet Loss~\eqref{eq:triplet} and Triplet Loss\_3~\eqref{eq:triplet_improved} still achieve and average clustering accuracy of 75.0\%, respectively, even though 20\% of wrong samples are used for training.
A similar behavior can be observed in Figure~\ref{fig:acc_noise}, top right, where we evaluate the same embeddings using k-means clustering. 
The observations are slightly different when adding label noise to the negative pairs, Figure~\ref{fig:acc_noise}, second row. Even when introducing 10\% noise, which corresponds to drawing negative samples completely at random in a balanced 10 class classification problem, all loss variants are relatively robust, especially for k-means clustering (Figure~\ref{fig:acc_noise}, second row, right). Yet, the proposed Triplet Loss\_3 again performs best.
In the bottom row of Figure~\ref{fig:acc_noise}, we consider an equal amount of noise on both positive and negative samples (corresponding to the diagonal in Table~\ref{tab:table1}). In this setting, the proposed Triplet Loss\_3 again shows higher stability than the two previous variants when clustering using minimum cost multicuts Figure~\ref{fig:acc_noise}(bottom left). For k-means clustering, the improvement over the Triplet Loss~\cite{schroff2015facenet} is marginal. This result is actually expected: The traditional Triplet Loss~\cite{schroff2015facenet} creates an embedding such that, for every data point, data points from the same class are closer than points from any other class. This fits well with the k-means clustering objective, assigning every points to the nearest cluster center, regardless their absolute distance. Yet, it can be problematic in the context of correlation clustering, where the minimum absolute distance between two clusters matters (compare again Figure~\ref{fig:Fig1}).

%\iffalse

%\fi  % Noise

\subsection{Qualitative Results}
\label{subsec:qualitative}

% -----------------------------------------

Figure~\ref{fig:tsne} shows a TSNE-visualization~\cite{maaten2008visualizing} of the embedding features learned from the CNN-model using the Triplet Loss\_3~\eqref{eq:triplet_improved} variant.
We use the minimum cost multicut approach to cluster CIFAR-10 test dataset.
In the particular experiment example, the total number of clusters are 44 with a cluster accuracy of 80.27\%.
The different colors represent the found class labels while in the ground truth, there are only 10 classes on the CIFAR10 dataset (which is shown in the legend).
Any other found clusters are considered as false positives and thus lower the cluster accuracy. 
However, there are in fact 34 small clusters that contain less than 10 images.
Three examples of such mini clusters are shown in bottom left and right as well as on the top left corner.
Even though there are false positives shown in the examples of the smaller clusters, the multicut approach explores meaningful sub-clusters within a class label, which may be desirable on real-world scenarios.
For instance, instead of finding the class \textit{horse} (in cyan), a subclass \textit{white-horses} is also found.
 % qualitative

%-----------------------------------------------
%%%%%%%%% BODY TEXT
\section{Conclusion}
\label{sec:conclusion}
In this work, we presented an extensive study on three different variations of the Triplet Loss.
Specifically, we have studied the clustering behavior of \textit{k-means} and minimum cost multicut clustering, applied to learnt embedding spaces from three Triplet Loss formulations on the CIFAR10~\cite{krizhevsky2009learning} dataset under a varying amount of label noise. %graph-based clustering approach based on learned features from a convolutional neural network (CNN).
We find that, while the traditional Triplet Loss~\cite{schroff2015facenet} is well suited for \textit{k-means} clustering, its performance drops under the looser assumptions made by minimum cost multicuts. We proposed a simplification of the Triplet Loss from \cite{zhang2016deep}, which allows to directly compute the probability of two data points for belonging to disjoint components. In a line of experiments on the CIFAR-10 dataset, we show that this proposed loss is robust against label noise in both clustering scenarios and outperformes both previous Triplet Loss versions in terms of clustering performance and stability.
\newpage

\bibliographystyle{IEEEtran}
\bibliography{paper}
\end{document}